\documentclass{article}

\usepackage[utf8]{inputenc} 
\usepackage[T1]{fontenc}    
\usepackage{hyperref}       
\usepackage{url}            
\usepackage{booktabs}       
\usepackage{amsfonts}       
\usepackage{nicefrac}       
\usepackage{microtype}      
\usepackage{lipsum}			
\usepackage{graphicx}
\graphicspath{ {} }
\usepackage[numbers]{natbib}
\usepackage{doi}
\usepackage{caption}
\usepackage{subcaption}
\usepackage{authblk}
\usepackage{multirow}
\usepackage{float}
\usepackage[nobottomtitles*]{titlesec}
\usepackage{xcolor}
\usepackage{enumitem}
\usepackage{bbold}
\usepackage{amsmath}
\setlist{noitemsep}

\providecommand{\keywords}[1]
{
  \small
  \emph{\textit{Keywords---}} #1
}

\def\correspondingauthor{\footnote{
	Corrersponding author.
	\emph{Email address:} jalmari.tuominen@tuni.fi
}}

\title{Forecasting Emergency Department Crowding with Advanced Machine Learning Models and Multivariable Input}

\author[1]{Jalmari Tuominen\correspondingauthor{}}
\author[1]{Eetu Pulkkinen}
\author[2]{Jaakko Peltonen}
\author[2]{Juho Kanniainen}
\author[1,3]{Niku Oksala}
\author[1,5]{Ari Palomäki}
\author[1]{Antti Roine}

\affil[1]{\footnotesize Faculty of Medicine and Health Technology, Tampere University}
\affil[2]{\footnotesize Faculty of Information Technology and Communication Sciences, Tampere University}
\affil[3]{\footnotesize Centre for Vascular Surgery and Interventional Radiology, Tampere University Hospital}
\affil[5]{\footnotesize Kanta-Häme Central Hospital, Hämeenlinna, Finland}

\hypersetup{
pdftitle={A template for the arxiv style},
pdfsubject={q-bio.NC, q-bio.QM},
pdfauthor={David S.~Hippocampus, Elias D.~Striatum},
pdfkeywords={First keyword, Second keyword, More},
}

\begin{document}
\maketitle

\AtBeginEnvironment{tabular}{\small}
\AtBeginEnvironment{caption}{\normal}

\begin{abstract}

Emergency department (ED) crowding is a significant threat to patient safety and it has been repeatedly associated with increased mortality. Forecasting future service demand has the potential patient outcomes. Despite active research on the subject, several gaps remain: 1) proposed forecasting models have become outdated due to quick influx of advanced machine learning models (ML), 2) amount of multivariable input data has been limited and 3) discrete performance metrics have been rarely reported. In this study, we document the performance of a set of advanced ML models in forecasting ED occupancy 24 hours ahead. We use electronic health record data from a large, combined ED with an extensive set of explanatory variables, including the availability of beds in catchment area hospitals, traffic data from local observation stations, weather variables, etc. We show that N-BEATS and LightGBM outpeform benchmarks with 11 \% and 9 \% respective improvements and that DeepAR predicts next day crowding with an AUC of 0.76 (95 \% CI 0.69-0.84). To the best of our knowledge, this is the first study to document the superiority of LightGBM and N-BEATS over statistical benchmarks in the context of ED forecasting.

\keywords{Emergency department, Crowding, Overcrowding, Forecasting, Multivariable analysis}

\end{abstract}


\section{Introduction}\label{introduction}
Emergency department (ED) crowding is a well-known threat to patient safety \cite{Boyle2012} and the documented adverse effects range from a decrease in the work satisfaction of ED staff \cite{Eriksson2018} to increased length of stay \cite{McCarthy2009} and increased mortality \cite{Guttmann2011, Richardson2006, Berg2019, Jo2014}. In contrast to outpatient clinics or elective surgery, EDs are unable to adjust the inflow of patients and are thus exposed to both a stochastic incidence of diseases and changes in patients’ care-seeking behaviour. Moreover, EDs are also usually unable to freely adjust the outflow of patients, since they depend on other health care facilities to organize follow-up care when necessary. The only component that an ED can independently adjust is throughput, which is mostly affected by the quantity \cite{Bucheli2004} and quality \cite{Trotzky2021} of staff. These restrictions lead to repeated crowding. Successful forecasts of future service demand would enable proactive administrative decisions aiming to alleviate or even prevent crowding, and have the potential to improve patient outcomes.  This rationale has motivated an increasing amount of ED forecasting articles \cite{Gul2018} but for some reason, readily available solutions have not emerged. We believe this to be, at least in part, due to three major gaps in current literature: 1) outdated forecasting methods, 2) lack of relevant multivariable input and 3) lack of binary performance metrics. 

First, a significant amount of ED forecasting literature has focused on the Autoregressive Integrated Moving Average (ARIMA) or it’s variants \cite{Gul2020}. The tendency to favour an ARIMA model over advanced models is understandable, as it has up to very recent years repeatedly defended its place over more complex solutions \cite{Zhou2018, Whitt2019, Cheng2021} and has been considered to be \emph{a pinnacle of statistical approach} in time series forecasting in general \cite{Oreshkin2019}. This has been changing rapidly. In 2020, a statistical and deep learning (DL) hybrid proposed by Smyl et al \cite{Smyl2020} outperformed statistical benchmark models in the renowned M4 time series forecasting competition for the first time in the history of the competition \cite{Makridakis2020}. Following this result, several time-series-specific DL architectures have been introduced, such as the Temporal Fusion Transformer (TFT) by Lim et al 2021 \cite{Lim2021}, Neural Expansion Analysis for Interpretable Time Series Forecasting (N-BEATS) by Oreshkin et al 2019 \cite{Oreshkin2019} and DeepAR by Salinas et al 2020 \cite{Salinas2020}. In the latest M5 competition in 2022, all of the best-performing models were pure ML implementations, and out of the five best performing solutions, four utilized a multivariable LightGBM model. In addition to LightGBM, DeepAR and N-BEATS were highlighted for \emph{showing forecasting potential}. However, these models have not been tested using ED data.

Second, the amount and quality of used input data has been limited. This is an important deficit because of the highly interdependent nature of the ED. As suggested by Asplin et al \cite{Asplin2003}, ED crowding is a sum of three operational components: input (number of arrivals), throughput (length of stay mainly affected by staffing resource) and output (mainly affected by availability of follow-up care beds). A disturbance in one of these components alone can lead to crowding, but the most severe situations are observed when two or more of them are detrimentally affected. This is in line with findings of M5, in which one of the seven key implications of the competition was the importance of exogenous variables \cite{Makridakis2022}. Despite this interdependent nature, ED forecasting input data has repeatedly consisted of calendar and weather variables, mounting up to 29 input vectors in total \cite{Whitt2019, Holleman1996, Jiang2018}. The studies that have suggested the utility of novel input variables, such as website visits \cite{Ekstrom2015}, road traffic flow \cite{Rauch2019} or the emergency department severity index \cite{Cheng2021}, have done so by utilizing only one of them at a time, which runs the risk of an overoptimistic evaluation of variable importance due to the inevitable multicollinearity between them. We thus believe that a data-centric approach with a high number of input variables has the potential to increase predictive accuracy, provide a better understanding of the factors underlying crowding and even inform policies among local health care providers. This is a continuation of our previous work, in which we used simulated annealing and floating search to perform feature selection in order to enhance accuracy, using conventional statistical models \cite{Tuominen2022}.

 Third, proposed ED forecasting models have been predominantly assessed using continuous error metrics such as mean absolute percentage error (MAPE) \cite{Gul2018}. This is a problem because these metrics are difficult to communicate to ED administration who are more familiar with binary metrics widely used in diagnostics, such as the area under the receiver opearting characteristics curve (AUROC) which is only rarely reported \cite{Hoot2009, Holleman1996}. Continuous metrics are even more difficult to use in  operative decisionmaking, since reactions to crowding are always inevitably discrete in nature. Moreover, many of the studies that have documented an association between mortality and crowding, have done so by comparing the most crowded quartile with less crowded ones \cite{Richardson2006, Berg2019, Jo2014}. We believe this kind of mortality-associated crowding to be the ultimate target for any forecasting model as we have previously argumented \cite{Tuominen2023} and since that target is by definition binary, we believe that metrics should be too.

To conclude, our contributions are as follows: 1) we investigate the performance of state-of-the art ML models in predicting ED occupancy using data spanning over 2 years in a large, combined ED; 2) we use the largest-to-date collection of explanatory variables, containing not only weather and calendar variables, but also the availability of hospital beds, traffic information, local public events, website visits and more, and analyse the proportional importance of these variables; and 3) we evaluate the models not only in continuous but also in binary terms using metrics that are easily understandable to ED stakeholders.

\section{Materials and Methods}\label{materials_and_methods}
\subsection{Data}\label{subsec:data}

Tampere University Hospital is an academic hospital located in Tampere, Finland. It serves a population of 535,000 in the Pirkanmaa Hospital District and, as a tertiary hospital, an additional population of 365,700, providing level 1 trauma centre capabilities. The hospital ED, \emph{Acuta}, is a combined ED with a total capacity of 111–118 patients, with 70 beds (with an additional 7 beds as a reserve) and 41 seats for walk-in patients. Approximately 100,000 patients are treated annually. For this study, all registered ED visits were obtained from a hospital database created during the sample period from January 1, 2017 to June 19, 2019. All remote consultations and certifications of death without prior medical interventions, as well as hourly duplicates, were excluded.

The dataset was split into three separate sets: the training, validation and test set. To capture yearly seasonal patterns in both the training and test sets, we ensured that the data was split into sets spanning a period of 12 months. This approach allows us to incorporate the full range of seasonal fluctuations throughout the year. The training set contained data from January 1, 2017 to  December 31, 2017, the validation set from January 1, 2018 to June 19, 2018, and the test set from June 20, 2018 to June 19, 2019. The sizes of the resulting sets were thus 8,736 (41\%), 4,056 (19\%) and 8,736 (41\%), respectively. The training set was used in training the models and the validation set in hyperparameter tuning. All accuracy metrics reported in this study are out-of-sample accuracy metrics calculated on the dedicated test set. Predictions were made 24 hours ahead at 00:00 on the test set, resulting in 24 x 364 matrices for each model. All data was scaled to a range of $[0,1]$ prior to analysis.

\subsubsection{Explanatory variables}

For the purposes of this study, we collected 181 explanatory variables from multiple data sources with the goal of covering as much of the three components of Asplin’s model as possible. These variables as summarized in Table \ref{tab:explanatory_variables} and briefly introduced below.  All covariates are divided into two categories: 1) past covariates ($P$) and 2) future covariates ($F$). $P$ features refer to variables that are not known in the future (e.g. hospital bed capacity) in contrast to $F$ features which are always known both in the future and in the past (e.g. hour of day). 

\paragraph{Population} The catchment area population was provided by Statistics Finland, since a linear association between catchment area size and service demand can be assumed. 

\paragraph{Traffic} Hourly traffic data were obtained from an open database maintained by Fintraffic Ltd, which is a company operating under the ownership and steering of the Finnish Ministry of Transport and Communications \cite{Digitraffic}. Data from all 33 bidirectional observation stations in the Pirkanmaa Region were included, resulting in 66 traffic feature vectors, each containing the number of cars that passed the observation station each hour. The acquisition of traffic variables was motivated by the work by Rauch et al \cite{Rauch2019}, which suggested that traffic variables might increase predictive accuracy when used as an input in an ARIMAX model. Locations of the observation stations along with their distance from study hospital are provided in Appendix \ref{appendix_c}.

\paragraph{Weather} Ten historical weather variables were collected from the nearest observation station located in Härmälä, Tampere, 600 meters from the city centre, using open data provided by the Finnish Meteorological Institute \cite{FMI}. The inclusion of weather variables was inspired by the work by Whitt et al \cite{Whitt2019}. We assumed that weather can be forecasted with satisfying accuracy one day in advance and, for this reason, used next-day weather variables as future covariates. 

\paragraph{Public events} City of Tampere officials provided us with an exhaustive historical event calendar, containing all public events ranging from small to large gatherings that were organized during the sample period in the Tampere area. We included three continuous public event variables: number of all events, number of minor public events and number of all major public events organized each day. Hypothetically, an increased number of citizens engaging in festivities – which often comes with increased substance consumption – might be associated with ED service demand.

\paragraph{Hospital beds} The temporal availability of hospital beds in 24 individual hospitals or health care centres in the catchment area was included as provided by the patient logistics system Uoma\textsuperscript{\textregistered} by Unitary Healthcare Ltd. in hourly resolution. In addition to the capacity of individual facilities, catchment-area-wide capacity was included as both the mean and sum of the capacity of the 24 facilities. The impact of the availability of beds on ED service demand is two-fold. First, a low availability of beds leads to a prolonged length of stay, since patients remain in the ED after initial treatment while they wait for an available follow-up care bed. This kind of access block leads to the cumulation of patients in the ED, and clinical experience has shown that this effect is a significant contributor to overcrowding. Second, a low availability of beds sometimes forces primary health care physicians to refer patients to an ED merely to organise the bed that the patient requires, which again contributes to crowding. Bed capacity statistics are visualised in Figure \ref{fig:beds} and locations of the facilities along with their distance from study hospital are provided in Appendix \ref{appendix_c}.

\begin{figure}[H]
	\centering
	\includegraphics[width=1.0\textwidth]{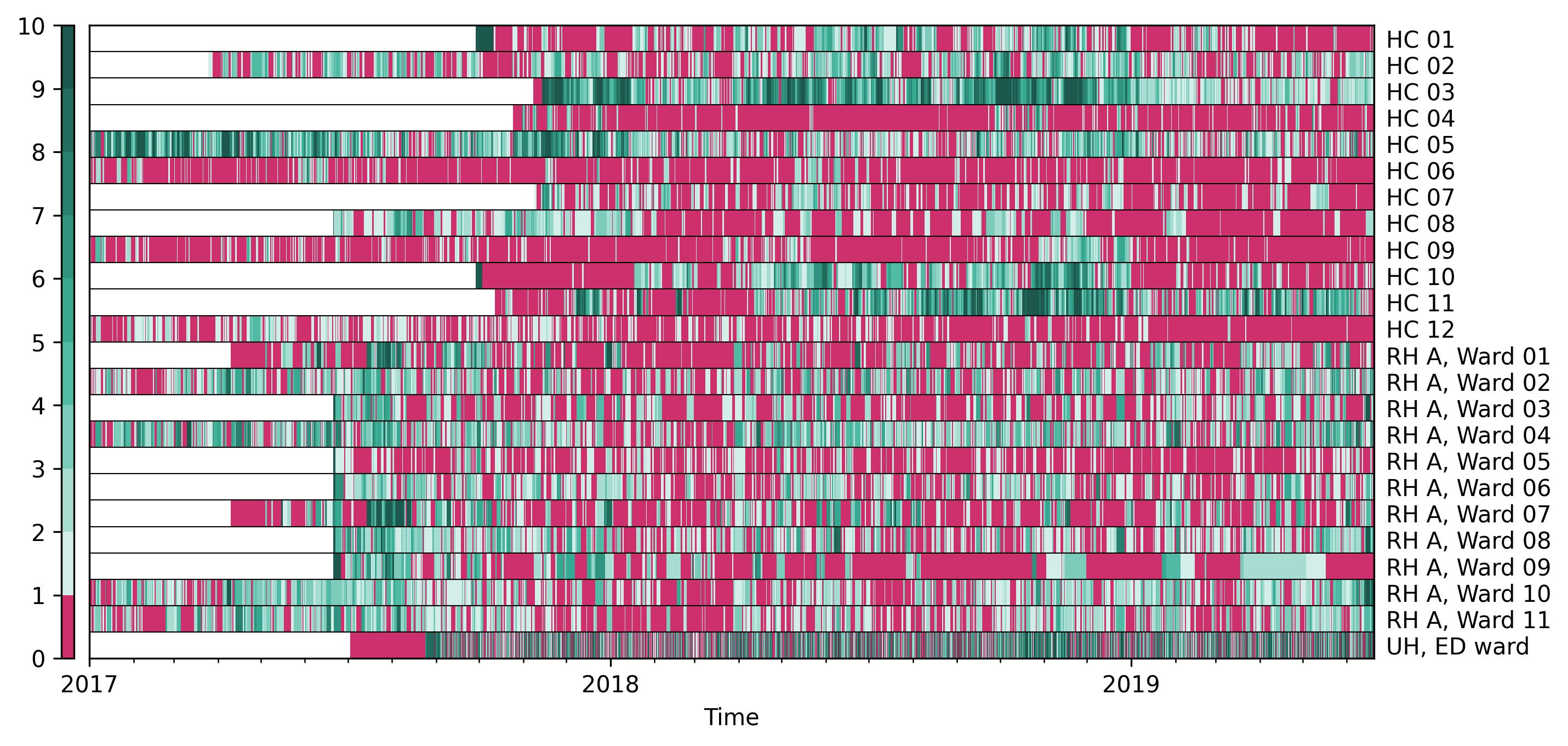}
	\caption{Hospital bed capacity statistics. HC = Health center, RH = Regional hospital, UH = University hospital. Red color indicates that no beds are available and number of available beds is shown with hues of green.}
	\label{fig:beds}
\end{figure}

\paragraph{Website visits} Data on website visits to two hospital domains were provided by the hospital IT department. Data were available on two domains: tays.fi (Domain 1, D1) and tays.fi/acuta (Domain 2, D2), the former of which being the hospital home page and the latter the home page of the hospital ED. D1 visit data were available in hourly resolution, whereas D2 data were only available in daily resolution. Using D1 visits, we also summed up visits between 6 p.m. and midnight in an identical manner to the one proposed by Ekström et al \cite{Ekstrom2015} ($\text{Domain 1}_{EV}$). In addition, we included a stationary version of this variable by dividing evening visits by earlier visits during the day ($\text{Domain 1}_{ER}$). The numbers of Google searches for the search term \emph{Acuta} were also extracted from Google Trends \cite{GoogleTrends}.

\paragraph{Calendar variables} The days of the month, weekdays and months were included as categorical variables. Timestamps of national holidays were provided by the University Almanac Office \cite{AlmanacOffice}, and each of them was included as a binary vector. Inspired by Whitt et al., we included “holiday lags”, which encode whether the two previous or two following days were holidays. We also included the numbers of previous consecutive holidays, encoding how many consecutive holidays preceded the day of interest. A binary encoding of a day’s status as a working or non-working day was also included.

\paragraph{Technical analysis} In addition to the exogenous explanatory features described above, we engineered 30 features using the endogenous signal of the target variables. These variables range from a set of moving averages and mathematical moments to econometric indicators and they are introduced in detail in Appendix \ref{appendix_b}.

\begin{table}[H]
\centering
	\caption{Explanatory variable list. $P$ = past covariate i.e. a value that is not known into the future at prediciton time, $F$ = future covariate i.e. a value that is known both in the past and in the future. *Some variables are provided elsewhere for brevity: Locations, names and distances of both hospital bed and traffic variables are provided in Appendix \ref{appendix_c}. TA indicators are listed and described in Appendix \ref{appendix_b}.}
	\label{tab:explanatory_variables}
	\begin{tabular}[t]{ llll }
	\hline
	Feature group & Name & Number \\
	\hline
	Traffic & * & $P_{1-66}$ \\
	Hospital beds & * & $P_{67-82}$ \\
	Website visits & Domain 1 & $P_{83}$ \\
	& $\text{Domain 1}_{EV}$ & $P_{84}$ \\
	& $\text{Domain 1}_{ER}$ & $P_{85}$ \\
	& Domain 2 & $P_{86}$ \\
	& Google Trends & $P_{87}$ \\
	TA indicators & * & $P_{88-117}$ \\
	Weather & Cloud count & $F_{1}$ \\
	& Air pressure &  $F_{2}$ \\
	& Relative humidity & $F_{3}$ \\
	& Rain intensity & $F_{4}$ \\
	& Snow depth & $F_{5}$ \\
	& Air temperature & $F_{6}$ \\
	& Dew point temperature & $F_{7}$ \\
	& Visibility & $F_{8}$ \\
	& Air temp min & $F_{9}$ \\
	& Air temp max & $F_{10}$ \\	
	Calendar variables & Holiday lags & $F_{11-18}$ \\
	& Working day & $F_{19}$ \\
	& Preceeding holidays & $F_{20}$ \\
	& National holidays & $F_{21}$ \\
	& Hours & $F_{22-44}$ \\
	& Weekdays & $F_{45-50}$ \\
	& Months & $F_{51-61}$ \\
	Public events & All events & $F_{62}$ \\
	& Minor events & $F_{63}$ \\
	& Major events & $F_{64}$ \\
	\hline
	& Total & $A_{181}$ \\
	\hline
	\end{tabular}
\end{table}

\subsubsection{Feature sets}

Using a high number of input features from multiple data sources poses a significant challenge if the predictive model is to be implemented in a real-life clinical setting, which increases both the cost of building and maintaining the system as well as its fragility. For this reason, we tested the models with two sets of inputs, one containing all variables listed above (feature set $A, n = 181$) and one containing  nothing but the target variable history (feature set $U$) as an input. Each multivariable model is tested with both $A$ and $U$ inputs, and they are distinguished from one another with naming convention of {$M_F$}, in which $M$ stands for model name and $F$ for feature set. For example, $\text{LGBM}_A$ refers to a LightGBM model trained and tested with all available data. We note that because DeepAR accepts only inputs that are known into the future, the multivariable version of the model uses only \emph{future covariates} as listed in Table \ref{tab:explanatory_variables}.

\subsubsection{Target variables}\label{sec:dependent_variables}

In this study, we have two target variables: 1) hourly absolute occupancy (HAO) and 2) binary crowded state (BCS).

\paragraph{Hourly absolute occupancy (HAO)} In this study we focus on predicting absolute nonstratified hourly occupancy of the ED. This includes both bed-occupying and walk-in patients in all treatment spaces of the ED. Occupancy was selected as the target variable for several reasons. First, it is affected by all three components of the Asplin's model, since input, throughput and output all contribute to total occupancy contrary to arrivals, which is by definition affected primarily by input. Second, occupancy is intuitively associated with crowding and it is easy to transform into binary crowded state as explained below.

\paragraph{Binary crowded state (BCS)} The word \emph{crowding} implies two possible states of the ED: normal operating conditions and a crowded state. Although this is intuitively obivious, an international and widely accepted definition for crowded state does not exist. Some crowding frameworks such as NEDOC \cite{Weiss2004} or EDWIN \cite{Bernstein2003} have been proposed but they have not been widely adopted and are not applicable in this study because of the variables they require are not available in the dataset.

In Tampere University Hospital, the ED is considered crowded whenever the absolute occupancy exceeds 80 patients. Exceeding this thresholds triggers a crowding protocol that allows the shift-supervising physician to call-in additional staff and mandates follow-up care facilities to receive patients even if their nominal capacity has been exceeded. This threshold annotates the top 4 \% of hours as crowded and these crowded hours are observed on 26 \% of all days as later described in more detail in Section \ref{results}. This aligns relatively well with work by Richardson et al in which the most crowded 25 \% of shifts were associated with increased mortality \cite{Richardson2006}. 

Hence, in this study, all hours $t$ where $y_t \ge 80$ were considered crowded. We evaluate the models ability to predict if the ED will reach BCS over the course of next 24 hours. Since predictions are always made at midnight as described in subsection \ref{subsec:data}, we consequently assess if the models can predict if the next day will be crowded or not.

\subsubsection{Performance metrics} 

\paragraph{Continuous metrics } We provide three continuous error metrics: Mean absolute error (MAE) and Root mean squared error (RMSE) for point forecasts (PF) and Mean scaled interval score (MSIS) for prediction intervals. MAE is calculated as follows:

\begin{equation}
	\mathrm{MAE} = \frac{1}{n} \sum_{t=1}^{n} |y_t - \hat{y}_t|,
\end{equation}
where $y_t$ is the ground truth and $\hat{y}$ is the predicion. MAE was used to calculate the proportional difference between the models and relevant benchmark and for statistical tests.

Point forecasts were also evaluated using root mean square error (RMSE):

\begin{equation}
\mathrm{RMSE} = \sqrt{\frac{1}{n}\sum_{t=1}^{n}(y_t-\hat{y}_t)^2}
\end{equation}

All the models investigated in this study are probabilistic in nature and were configured to produce 95 \% prediction intervals (PI) in addition to PFs. We thus also quantify the performance of these PIs using MSIS as proposed by Gneiting and Raftery \cite{Gneiting2007}:

\begin{equation}
\mathrm{MSIS} = \frac{\Sigma_{t=n+1}^{n+h}(u_{t}-l_{t})+\frac{2}{\alpha}\mathbb{1}\{y_{t}<l_{t}\}+\frac{2}{\alpha}(y_{t}-l_{t})\mathbb{1}\{y_{t}>u_{t}\}}{h \times \frac{1}{n-m}\Sigma_{t=m+1}^{n}|y_{t}-y_{t-m}|}
\end{equation}
where $u_t$ and $l_t$ are the upper and lower bounds, $x_t$ is the ground truth and $\alpha$ is the significance level which was set to 0.05 based on the levels of the generated PIs.

\paragraph{Binary metrics} Additionally, area under the receiver-operating characteristic curve (AUROC) was calculated for each model in predicting next-day crowding, using the definition provided in Section \ref{sec:dependent_variables}. The 95\% confidence interval for the area under the curve (AUC) was calculated using bootstrapping with 250 iterations. Due to class imbalance between crowded an non-crowded states, the overrepresented class was randomly downsampled to match the underrepresented ones.

\paragraph{Statistical significance} Statistical significance testing of hourly absolute error rates between ARIMA and other models was performed using Kruskal-Wallis one-way analysis for variance with Dunn's post hoc test and Holm's correction for multiple pairwise comparisons. This aims to provide a similar approach as in M3 \cite{Koning2005}. The significance level was specified as p < 0.05. Statistical tests were performed using Scipy \cite{Scipy} and scikit-posthocs \cite{Terpilowski2019}.

\subsubsection{Feature importance analysis}

Studying how the model selects and weights features underlying the predictions can provide insight into: 1) factors behind crowding and; 2) reasons underlying good or bad forecasting performance. In this study, this was performed using SHapley Additive exPlanations (SHAP) method as proposed by Lundber et al 2017 \cite{Lundberg2017}. SHAP assigns a unique importance value to each feature in a prediction by quantifying its contribution to the prediction outcome. SHAP values are based on cooperative game theory principles, calculating the average marginal contribution of each feature across different coalitions of features, providing a unified and interpretable explanation for individual predictions. For brevity, we limit our attention to importance statistics of the $\text{LightGBM}_A$ model.

\subsection{Models}

In this study we document performance of four (4) forecasting models: LightGBM, Temporal Fusion Transformer (TFT), DeepAR, and N-BEATS benchmarked against four (4) benchmarking models. Model definition, training and backtesting was handled using software by Herzen et al \cite{Herzen2022} which provided a unified interface to underlying models and also provided an implementation unless otherwise stated. Hyperparameter optimisation (HPO) is performed for all the models using Tree-structured Parzen Estimator (TPE) \cite{Bergstra2011} with optimization framework by Akiba et al \cite{Akiba2019}. Detailed HPO procedure and search spaces are provided in the Appendix \ref{appendix_a}. Below, a short description of each model is provided.

\paragraph{Benchmark models} Four models were used for benchmarking purposes: Seasonal naïve (SN), Autoregressive Integrated Moving Average (ARIMA), and two ETS models: Holt-Winter's Seasonal Damped method (HWDM) and Holt-Winter's Seasonal Additive Method (HWAM). 168-hour sliding window was utilized for all models. ARIMA parameters were defined with AutoARIMA as initially described by Hyndman et al \cite{Hyndman2008} using the stepwise approach and Python implementation by Garza et al \cite{garza2022statsforecast}. \textit{A priori} known hourly seasonality 24 was provided to AutoARIMA model as a parameter.

\paragraph{Temporal fusion transformer}
Transformer models have quickly gained popularity ever since their introduction \cite{Vaswani2017}, especially in the field of natural language processing (NLP). Recently, these models have shaken the world by introduction of large language models, most notably Generative Pre-Trained Transformer (GPT) \cite{OpenAI2023}. The temporal fusion transformer (TFT) was introduced by Lim et al, with the goal of bringing the benefits of transformer models to the domain of time series forecasting \cite{Lim2021}. The TFT model also employs an LSTM encoder–decoder structure to generate a set of uniform temporal features, which serve as input to the temporal self-attention layer. This structure allows the TFT to pick up long-range dependencies that are often challenging for RNN-based architectures to learn. A gating mechanism implemented with the gated residual network (GRN) allows the model to adapt to different datasets and scenarios by limiting the amount of nonlinearity applied in different parts of the architecture. This is especially useful with small or noisy datasets, in which case simpler models are known to provide better results. Variable selection networks allow the model to learn which inputs are relevant at each timestep. Through these, the TFT can remove any unnecessary noisy inputs which could negatively impact performance. In turn, more weight can be given to the most salient covariates, improving model performance. Static covariate encoders are wired into various locations of the TFT network, aiming to provide additional information and context to the model. Multi-head attention improves the capacity of a standard attention mechanism by employing different heads for different representation subspaces. This means that each head can learn different temporal patterns from the common set of inputs. The Python implementation of TFT by Beitner et al 2022 \cite{beitner2020pytorch} was used.

\paragraph{N-BEATS} Neural basis expansion analysis for interpretable time series forecasting (N-BEATS) is a univariable deep learning architecture introduced in 2020 by Oreshkin et al \cite{Oreshkin2019}. The N-BEATS architecture consists of blocks. Each block is a multi-layer, fully connected neural network, which takes the previous block’s input and produces two outputs, a forward forecast and a forecast in reverse time, i.e. a backcast. The backcast is used to create the next input by subtracting the backcast from the previous input. This helps the next block to make a better forecast by removing components of their input that are not helpful for forecasting. Each individual block forecast is added together to create the global model output forecast for the horizon. Individual blocks are grouped together in stacks, in which each stack can be given a specific task (e.g., trend stack or seasonality stack). In each stack, blocks share learnable parameters between each other for better performance.

\paragraph{DeepAR} DeepAR is a probalisitc autoregressive recurrent network model proposed by Salinas et al \cite{Salinas2020}. It enables transfer learning meaning that a single "global model" can be trained from multiple related time series and  utilizes a deep architecture with multiple layers of Long Short-Term Memory (LSTM) cells to capture both short-term dependencies and long-term patterns in the data. To enable probabilistic forecasting, DeepAR uses a distributional approach, where it learns to predict the parameters of a predefined probability distribution (such as quantile regression as used in this study) that characterizes the uncertainty of future predictions. This allows for generating not just point estimates but also prediction intervals, enabling decision-making under uncertainty. It also allows covariates given their values are known in "prediction range". In this study, these variables are referred to as \textit{future covariates}.

\paragraph{LightGBM}
LightGBM (LGBM) is an extension to gradient boosted decision tree (GBDT) proposed by Ke et al \cite{Ke2017}. It introduces two novel compoments: 1) Gradient-based One-Side Sampling (GOSS) and 2) Exclusive Feature Budnling (EFB) both of which aim to enhance the efficiency of the training procedure with minimal impact on accuracy. In the initial benchmarks, LGBM training times were up to 20 times faster compared to conventional GBDT with only marginally worse accuracy. Being a GBDT, the model allows covariates to be used as an input and it is naturally resistant to missing data. These features make LGBM an interesting candidate in our problem setting. It has shown state-of-the art performance in M5 competition \cite{Makridakis2022} but to the best of our knowledge, it has not been benchmarked using ED data. 

\section{Results}\label{results}
\begin{figure}[H]
    \centering
    \includegraphics[width=\textwidth]{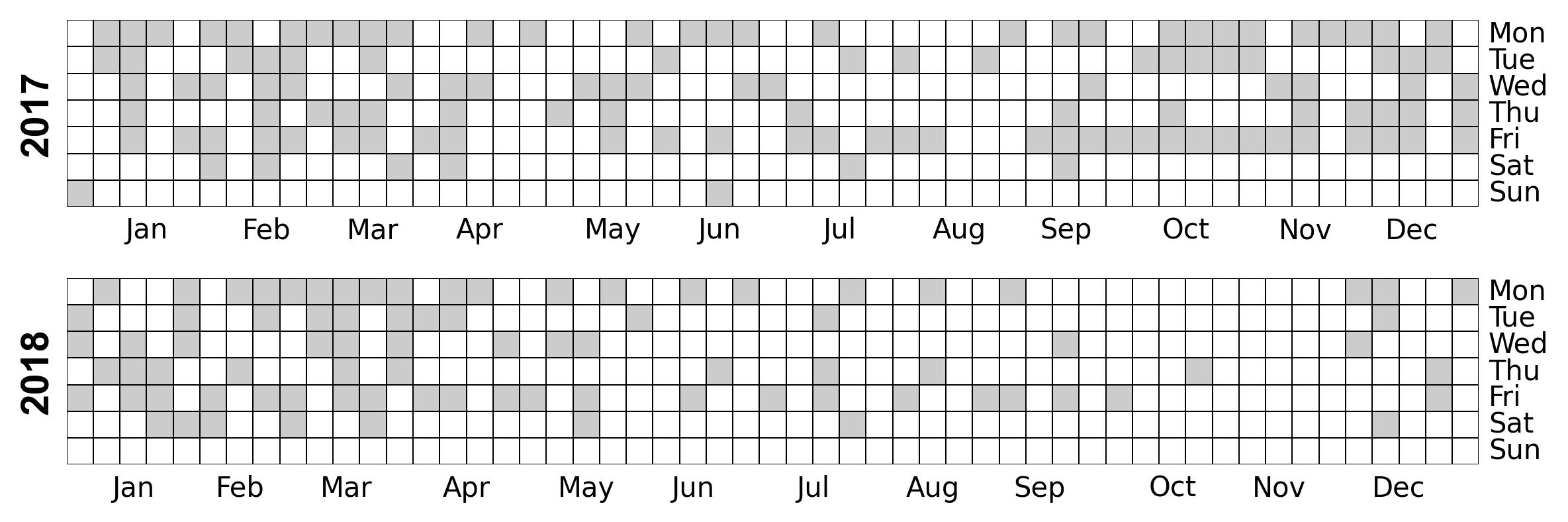}
    \caption{Calendar map of crowded days. All days during of which even one hour reached the binary crowded state are highlighted. Note the stochastic temporal distribution of the events with slight tendency of Mondays and Fridays to be crowded. The plot also shows the relative quietness of weekends in comparison to weekdays. Only years 2017 and 2018 are shown for brevity.}
    \label{fig:calmap}
\end{figure}

\paragraph{Descriptive statistics}\label{descriptive_statistics} The inclusion criteria resulted in a sample of 210,019 individual visits.  Minimum, median and maximum occupancies were 2, 38 and 124 respectively. BCS was observed on 792 (4 \%) out of the 21,600 hours in the sample. One or more hours were crowded on 235 (26 \%) days out of the 900 days in the sample. Temporal distribution of these days is provided in Figure \ref{fig:calmap}

\paragraph{Missing data} There was a significant amount of missing data in the case of available hospital beds, as can be seen in Figure \ref{fig:beds}. In total, data was missing on 77,636 (14\%) hours out of the total 561,600 hours for all facilities combined. The amount varied significantly between facilities from 0-7,486 hours (0-35\%) which is due to the gradual introduction of the Uoma\textsuperscript{\textregistered} software to each facility. All missing data were imputed with constant zero.

\subsection{Model performance}

\paragraph{Continuous performance} Continuous performance results are provided in Table \ref{tab:performance}. Kruskall-Wallis showed statistically significant differences between the models with p=0.0. $\text{N-BEATS}$ was the best-performing model in terms of MAE and RMSE with values of 6.98 and 87 respectively, yielding a 11 \% improvement over ARIMA (p<.001). $\text{LightGBM}_U$ was the second best model with MAE of 7.10 and proportional improvement of 9 \% over ARIMA (p<.001). $\text{TFT}_A$ outperformed ARIMA with MAE of 7.49 which yielded a 4 \% proportional improvement but the difference was not statistically significant (p=0.716). $\text{TFT}_U$ and $\text{DeepAR}_U$ with both feature sets were 2-10 \% worse than ARIMA but outperformed ETS models. Differences between ARIMA and $\text{TFT}_A$, $\text{DeepAR}_F$ and $\text{TFT}_U$ were not statistically significant. $\text{DeepAR}_U$ was significantly worse than ARIMA. N-BEATS and $\text{TFT}_U$ had the lowest MSIS of 47 compared to ARIMA's 65 and ETS models 55-56.

\begin{table}[H]
\centering
\caption{Continuous performance of the tested models.    FS = feature set, MAE = mean absolute error,     RMSE = root mean squared error,     MSIS = mean scaled interval score.     95 \% confidence intervals in parenthesis.}
\label{tab:performance}
\begin{tabular}{rcccccc}
\toprule
{} & FS &               MAE &  Delta \% &      p &           RMSE & MSIS \\
Model    &    &                   &          &        &                &      \\
\midrule
SN       &  U &  9.53 (9.40-9.68) &      -22 &  <.001 &  158 (153-164) &    - \\
HWAM     &  U &  9.44 (9.30-9.56) &      -21 &  <.001 &  149 (145-153) &   56 \\
HWDM     &  U &  9.20 (9.07-9.33) &      -18 &  <.001 &  142 (138-146) &   55 \\
DeepAR   &  U &  8.64 (8.53-8.79) &      -10 &  <.001 &  133 (129-137) &   55 \\
TFT      &  U &  7.94 (7.84-8.08) &       -2 &  1.000 &  114 (110-117) &   47 \\
ARIMA    &  U &  7.82 (7.70-7.93) &        0 &      - &  111 (107-114) &   65 \\
DeepAR   &  F &  7.75 (7.62-7.88) &        1 &  1.000 &  112 (108-117) &   75 \\
LightGBM &  A &  7.69 (7.57-7.80) &        2 &  1.000 &   101 (98-104) &   59 \\
TFT      &  A &  7.49 (7.37-7.61) &        4 &  0.921 &    98 (95-101) &   57 \\
LightGBM &  U &  7.10 (7.00-7.20) &        9 &  <.001 &     88 (85-90) &   57 \\
N-BEATS  &  U &  6.98 (6.86-7.09) &       11 &  <.001 &     87 (84-90) &   47 \\
\bottomrule
\end{tabular}
\end{table}

\begin{figure}[H]
    \centering
    \begin{subfigure}[b]{0.4\textwidth}
        \includegraphics[width=\textwidth]{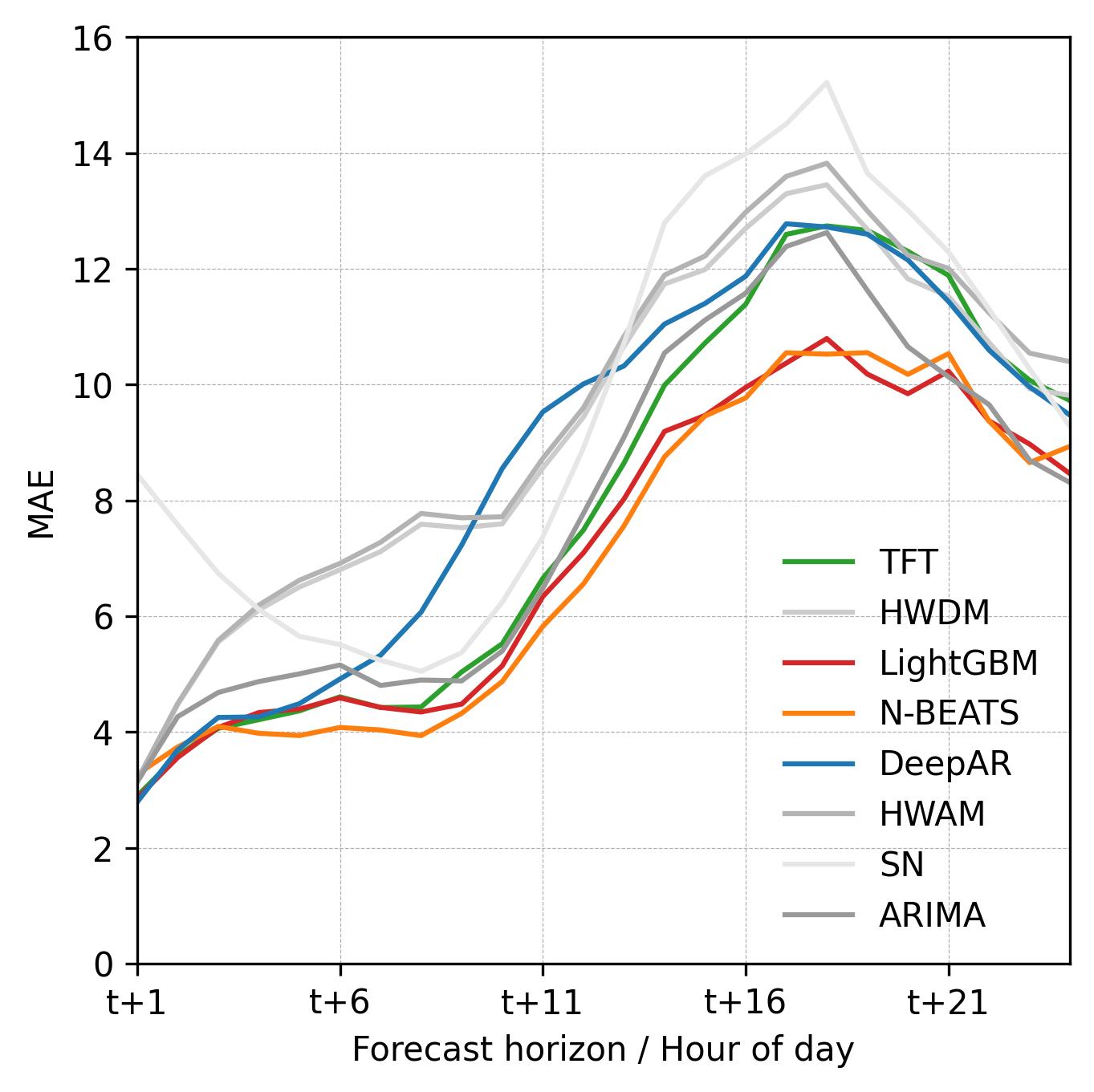}
        \caption{Univariable models}
        \label{fig:horizon_mae-u-1}
    \end{subfigure}
    \begin{subfigure}[b]{0.4\textwidth}
        \includegraphics[width=\textwidth]{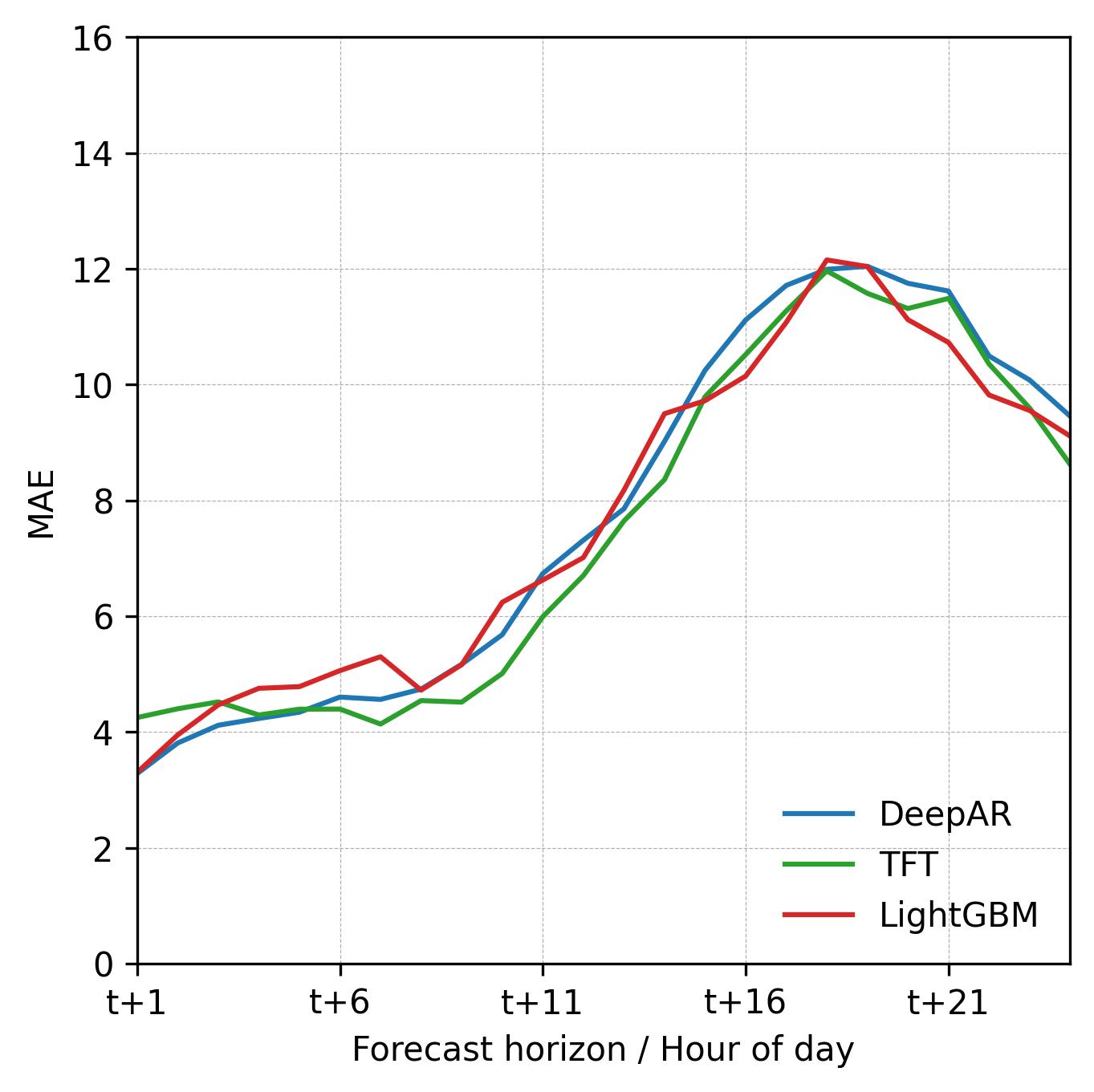}
        \caption{Multivariable models}
        \label{fig:horizon_mae-a-1}
    \end{subfigure}
    \caption{Horizontal error as measured by mean absolute error (MAE)}
    \label{fig:horizon_mae}
\end{figure}

\paragraph{Horizontal performance} The hourly accuracy of each model stratified by the forecasting horizon is provided in Figure \ref{fig:horizon_mae}. The figure show that variance between models is greatest in the afternoon which follows the increased variance of the target variable. $\text{LighGBM}_U$ and N-BEATS were consistently the best models regardless of the forecast horizons. Multivariable models performed in similiar manner.

\begin{figure}[H]
    \centering
    \begin{subfigure}[b]{0.4\textwidth}
        \includegraphics[width=\textwidth]{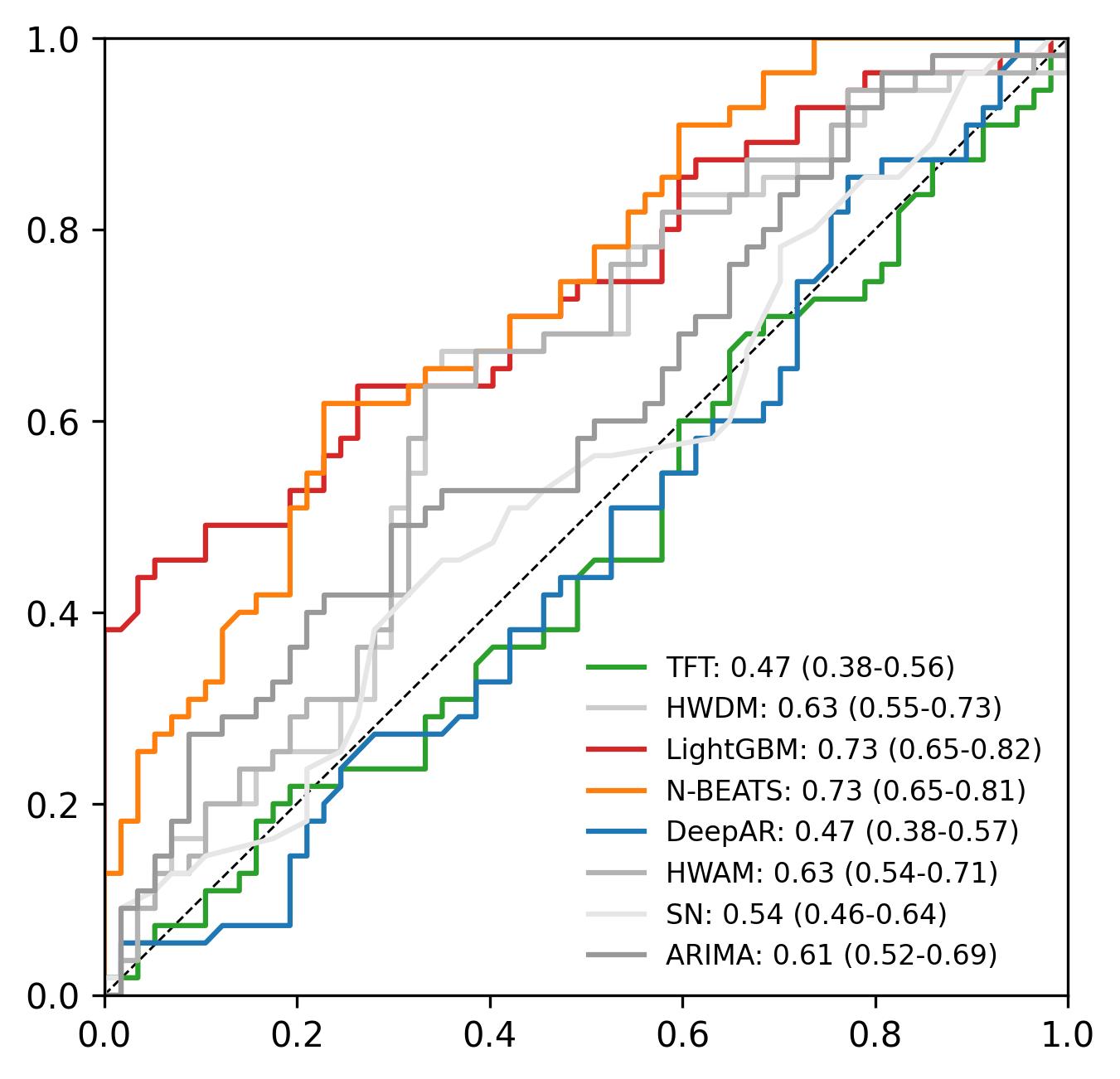}
        \caption{Univariable models}
        \label{fig:auc-u-1}
    \end{subfigure}
    \begin{subfigure}[b]{0.4\textwidth}
        \includegraphics[width=\textwidth]{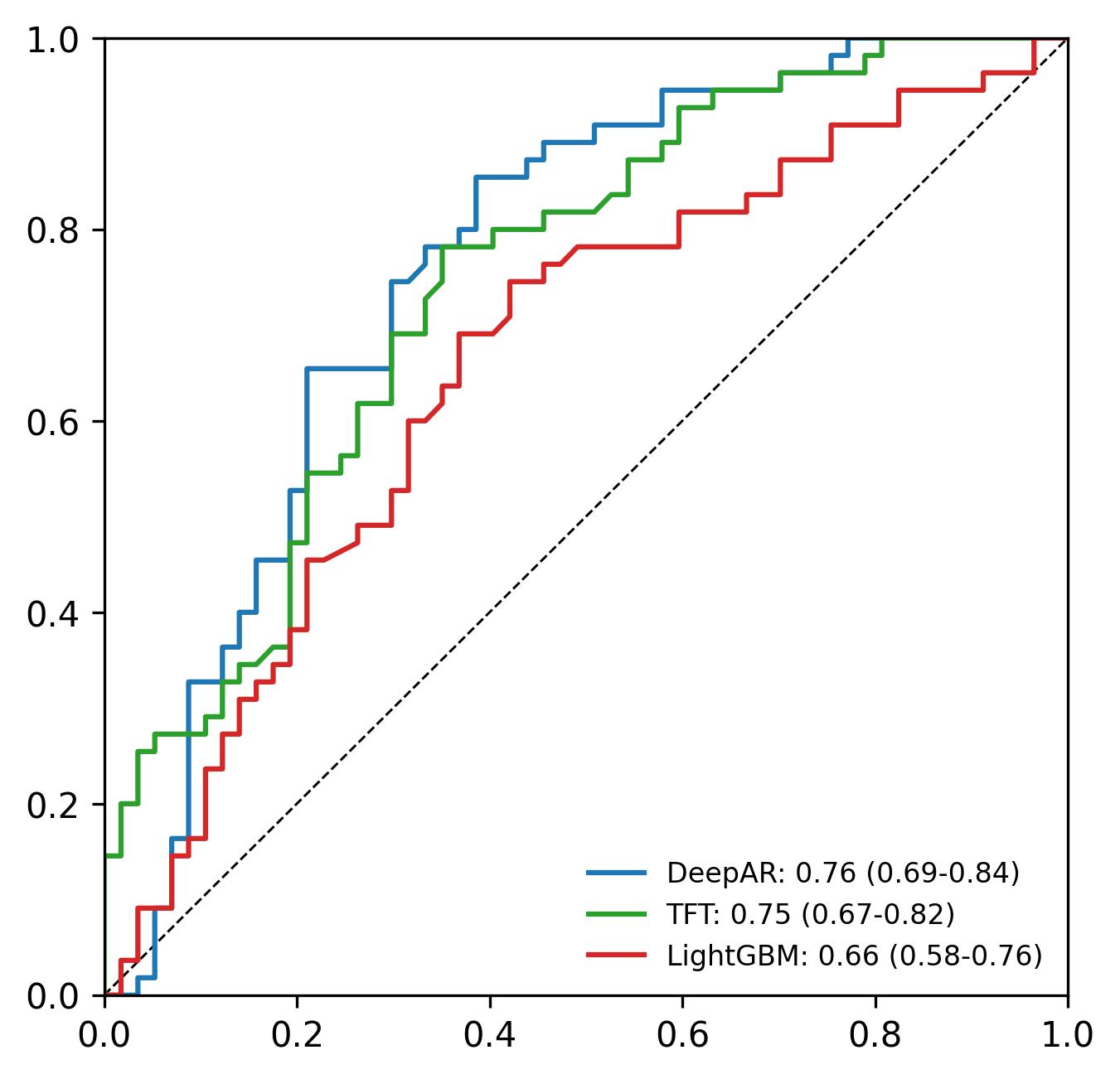}
        \caption{Multivariable models.}
        \label{fig:auc-a-1}
    \end{subfigure}
    \caption{Area under the receiver opreating characteristics curves (AUROC) in predicting next day crowded state.}
    \label{fig:auc}
\end{figure}

\paragraph{Binary performance} The receiver-operating characteristics of the models are provided in Figure \ref{fig:auc}. AUC values for benchmark models were 0.54, 0.60, 0.63 and 0.63 for SN, ARIMA, HWDM and HWAM respectively. $\text{DeepAR}_A$ was the best performing model with AUC of 0.76 (95\% CI 0.69-0.84) followed by $\text{TFT}_A$'s 0.75 (95\% CI 0.67-0.82). N-BEATS and LightGBM were the best univariable models: 0.73 (95\% CI 0.65-0.80) and 0.73 (95 \% CI 0.65-0.82) respectevely. $\text{TFT}_U$ and $\text{DeepAR}_U$ were both outpeformed by benchmarks.

\subsection{Feature importance analysis}
\begin{figure}[H]
    \centering
    \begin{subfigure}[b]{0.45\textwidth}
        \includegraphics[width=\textwidth]{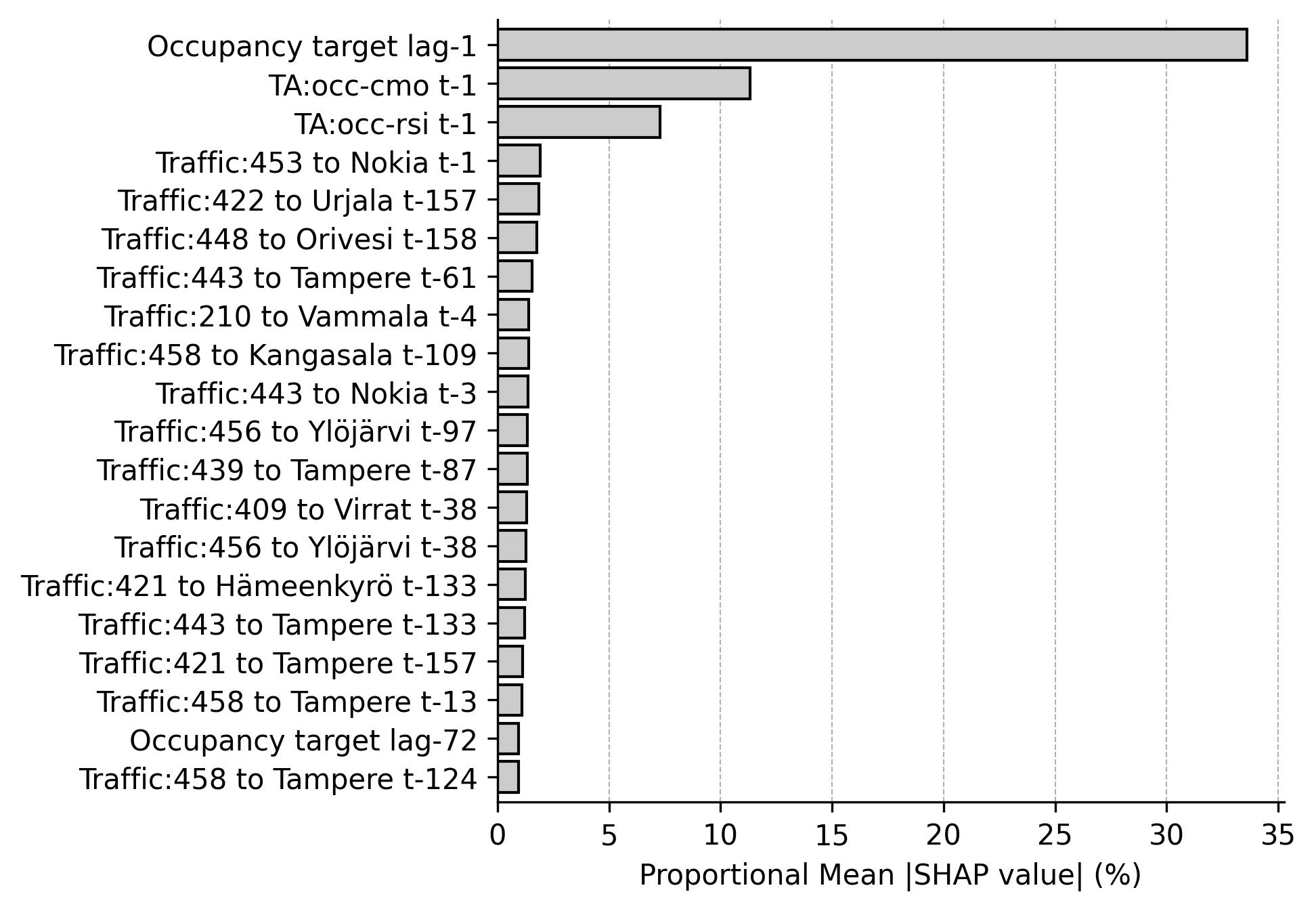}
        \caption{$t+1$ importance}
        \label{fig:importance-1}
    \end{subfigure}
    \begin{subfigure}[b]{0.45\textwidth}
        \includegraphics[width=\textwidth]{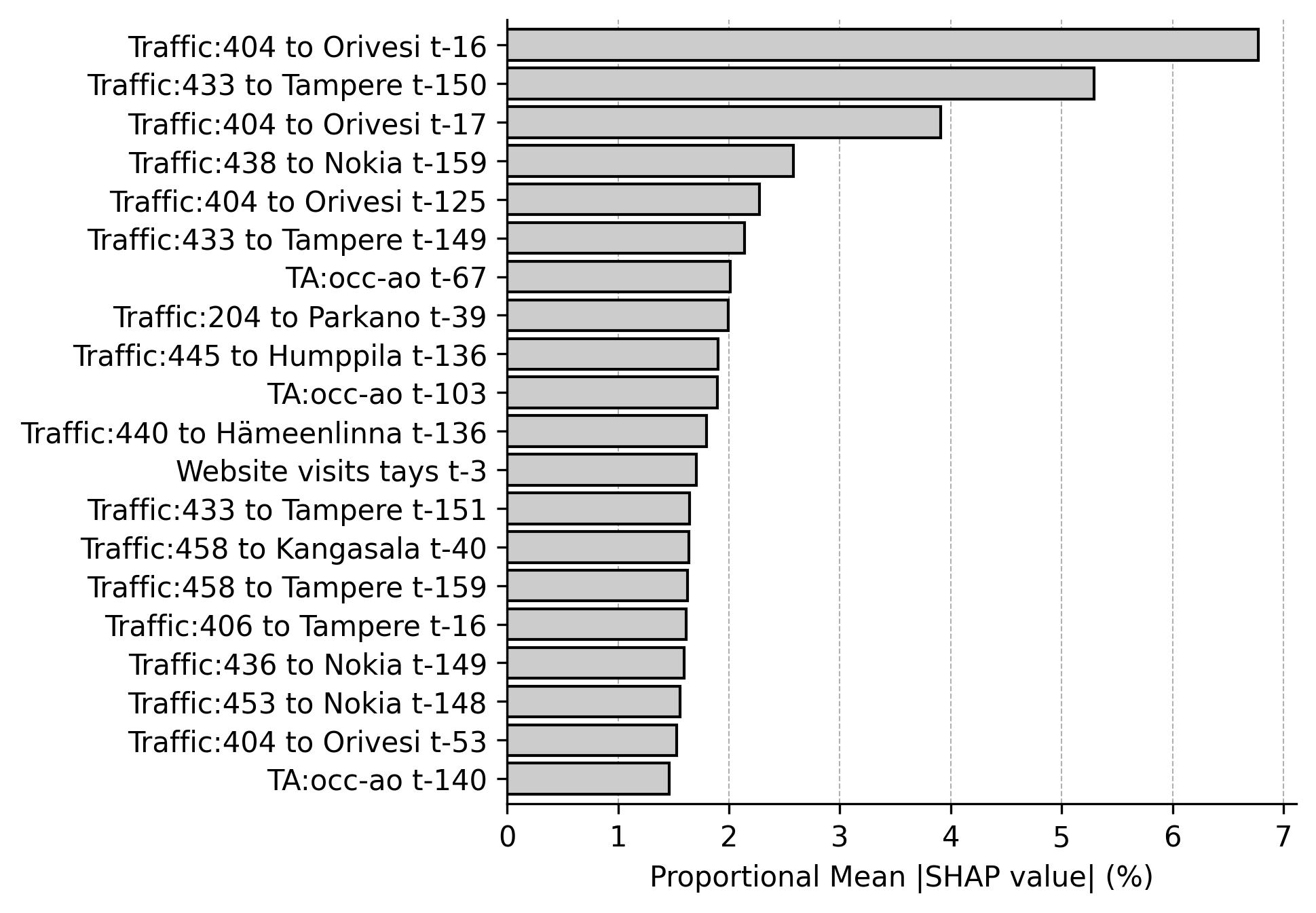}
        \caption{$t+24$ importance}
        \label{fig:importance-24}
    \end{subfigure}
    \caption{Feature importance SHAP statistics for 20 most important features used with $\text{LightGBM}_A$ model.}
    \label{fig:importance}
\end{figure}

Proportional SHAP values for 20 most important features for $\text{LightGBM}_A$ model are visualised in \ref{fig:importance} separately for horizons $t+1$ and $t+24$. For $t+1$, the target variable itself at lag $t-1$ was the most important variable followed by CMO and RSI and 16 traffic variables. Fort $t+24$ predictions, 16 traffic variables were included in top 20, along with website visit statistics and three different lags of the AO indicator

\section{Discussion}\label{discussion}
 
\subsection{Continuous performance} 
$\text{N-BEATS}$ was the best-performing model providing an 11 \% improvement over ARIMA. Coincidentally, Oreshkin reported an identical improvement of 11 \% over statistical models in their own benchmarks \cite{Oreshkin2019}. $\text{LightGBM}_U$ was the second best model providing a 9 \% improvement over ARIMA. This fell short of the 22.4 \% improvement gained the best-performing M5 solution that was also based on LightGBM. We are no aware of previous work that has benchmarked LightGBM in ED service demand forecasting. Gradient boosted machines on the other hand have been used by Caldas et al \cite{Caldas2022}. Based on their results, XGBoost was 16 \% worse than ARIMA which contrasts our results. We note that the second best solution in M5 competition was based on a N-BEATS and LightGBM hybrid. Findings of this study aligns with those of the M5 and validates the performance of these models using ED data.

$\text{TFT}_A$ was the third-best model with 4 \% improvement compared to ARIMA but the difference was not statistically significant. Using ED data, Caldas et al \cite{Caldas2022} reported a 5 \% improvement using multivariable TFT over ARIMA which aligns well with our results. TFT outperformed DeepAR, as demonstrated in the initial benchmarks of the original article by Lim et al \cite{Lim2021}. $\text{DeepAR}_F$ showed a 1 \% improvement over ARIMA but the difference was not statistically significant whereas $\text{DeepAR}_U$ was statistically significantly  worse than ARIMA by -10\%. This is in contrast with findings of M5 in which DeepAR performed markedly better \cite{Makridakis2022}. There are likely two main reasons for the suboptimal performance of DeepAR. First, the most important input variables belonged to $P$ group, as shown in Figure \ref{fig:importance} and DeepAR cannot account for these variables by design. Second, DeepAR has been specifically designed to be used in a setting where there are multiple similiar time series to learn from using transfer learning. Here, we used only one target variable as an input.

\subsection{Binary performance} 
Both $\text{LightGBM}_U$ and N-BEATS reached an acceptable AUC of 0.73 over ETS models' 0.63. This is encouraging and suggests that these models could indeed be utilized to aid administrative decision making in the near future. Univariable nature of these models is beneficial from practical standpoint, since they would be easy to incorporate to hospital IT infrastructure with minimal external software integrations. $\text{TFT}_A$ and $\text{DeepAR}_A$ provided a slightly better discrimnatory power with AUC's of 0.76 and 0.75 respectevily. 

Comparison of discrete performance with previous studies is difficult because directly comparable performance results have not been reported \cite{Gul2018}. Hoot et al \cite{Hoot2009} have documented the performance of a simulation based crowding software that was able to predict $t+8$ crowding with an AUC of 0.85. We have previously documented performance of a crowding early warning software that was able to predict afternoon crowding at 1 p.m. with an AUC of 0.84 using ETS models \cite{Tuominen2023}. We were unable to reach this performance when predicting next day crowding, which is likely due to significantly longer forecast horizon. The discrimatory power fell also short of some decision-making algorithms widely used in clinical practice, such as the National Early Warning Score for in-hospital mortality \cite{Loisa2022} (1-day AUC 0.91). Further work is required to improve the binary performance of the models. A natural next step would be to convert the propblem into classification task at the model training phase instead of the \emph{post hoc} approach adopted here.

\subsection{Feature importance analysis} 
Surprisingly, performance $\text{LightGBM}_A$ was markedly worse than $\text{LightGBM}_U$. We believe this to be due to overfitting especially as regards to traffic variables which were heavily weighted in SHAP statistics \ref{fig:importance}. We utilized a simple train-test split in this study and did not retrain the model after initial parameter estimation. It is thus possible that associations learned on the 2017 train set do not generalize over the year long test set. However, $\text{TFT}_A$ outperformed it's univariable counterpart $\text{TFT}_U$ and multivariable $\text{LightGBM}_A$. It is thus possible that TFT is more resistant to the abovementioned overfitting. Unfortunately, we were unable to produce feature importance statistics for the TFT model due to software limitations. 

The intuitive association between follow-up care capacity and increased occupancy was strong and we were expecting an improveded performance by including these variables. However, this was not observed, which might be due to a number of reasons. First, there was a significant amount of missing data in hospital bed variables, as described in section \ref{subsec:data} which might obscure their value. Second, it is possible that an unstratified sample played a role, since only 39\% of the presented patients were admitted after initial assessment, meaning that follow-up care capacity had no direct impact on the remaining 61\% of the patients. Third, the importance of these variables might become apparent if the models were refitted iteratively over the test set.

The included traffic monitoring nodes and their lags seem to have been selected arbitrarily. Moreover, the "importance" of traffic variables increases in proportion of forecast horizon as can be seen in Figure \ref{fig:importance}. We believe this to be due to overfitting as speculated above. This is in contrast with Rauch et al \cite{Rauch2019} who suggested that inclusion of these variables improve performance of an ARIMAX model. Website variables received only marginal weight and their inclusion did not improve performance which contradicts the findings of Ekström et al \cite{Ekstrom2015}.

\subsection{Limitations}
This study is limited by its retrospective single-centre setup, and further work is required to investigate the applicability of our approach to other facilities, preferably in a prospective setting. In the training set, we observed missing data on hospital beds variables which might obscure their importance. This study focused on unstratified sample, meaning that pooled occupancy statistics of all treatment rooms were used as the target variable. This is a limitation for two reasons. First, occupancies of different treatment spaces might have very different interactions with included explanatory variables. Second, unstratified sample does not allow us to leverage transfer learning capabilities e.g. DeepAR model which might improve it's performance.

\subsection{Future work}

We identify several directions for follow-up studies. It would be interesting to see whether the accuracy could be further improved by including triage variables, as suggested by Cheng et al \cite{Cheng2021}. Stratifying the total visit statistics by functional subunits of the ED (walk-in clinic, medical, surgical, etc.) would likely produce different feature importance statistics, potentially improve model performance and certainly be of operational value. Multivariable input did not provide marked performance improvement. It is possible that iterative parameter estimation at cross-validation stage could enable the models to adapt to inevitable temporal changes, which requires further investigation. Classification methods should be investigated in forecasting binary crowded state. Finally, the performance of the models have to be evaluated in a prospective setting.

\section{Conclusions}\label{conclusions}
In conclusion, this study suggests that in the ED forecasting context: 1) N-Beats and LightGBM models outperform both statistical benchmarks and other advanced ML models; 2) high-dimensional input data does not necessarily improve performance; and 3) it is possible to forecast next day ED crowding with an AUC of 0.76.

\bibliographystyle{unsrt}
\bibliography{references}  

\clearpage
\section*{Declarations}

\subsection*{Ethics approval and consent to participate}
Since our study was retrospective in nature, ethics committee approval was not required. An institutional approval was acquired prior to data collection, with the following specifications:
 
\begin{itemize}
    \item Name: Potilaslogistiikan häiriötekijöiden tunnistaminen ja mallintaminen
    \item Number: PSHP/R19565
    \item Date: 16 June 2019
\end{itemize}

\subsection*{Consent for publication}
Not applicable

\subsection*{Availability of data and materials}
The datasets used and/or analysed during the current study are not currently available due to legislative restrictions.

\subsection*{Conflict of interest statement}
NO is a shareholder in Unitary Healthcare Ltd., which has developed a patient logistics system currently used in the study emergency department. JT, AR and EP are shareholders in Aika Analytics Ltd., which is a company specialized in time series forecasting.

\subsection*{Funding}
The study was funded by the Finnish Ministry of Health and Social Welfare via the Medical Research Fund of Kanta-Häme Central Hospital, by the Finnish Medical Foundation, the Competitive State Research Financing of the Expert Responsibility Area of Tampere University Hospital, Pirkanmaa Hospital District, Grant 9X040 and Academy of Finland Grant 310617.

\subsection*{Authors’ contributions}
Study design (AR, JT, NO, AP, JP, JK). Data collection (JT, NO). Data-analysis (EP, JT). Technical supervision (JK, JP). Medical supervision (AR, NO, AP). Manuscript preparation (AR, JT, EP). All authors have read and approved the final manuscript.

\subsection*{Acknowledgements}
We acknowledge Unitary Healthcare Ltd for providing the dataset on available hospital beds, the City of Tampere for providing timestamps for public events, and Tampere University Hospital information management for providing website visit statistics. We also acknowledge \texttt{REDACTED PER REQUEST OF THE ASSOCIATE EDITOR} for providing computational resources and specifically D.Sc. Mats Sjöberg for technical support.

\clearpage
\appendix

\section{Appendix: Hyperparameter optimisation}\label{appendix_a}

\setcounter{table}{0}
\renewcommand{\thetable}{A\arabic{table}}
\setcounter{figure}{0}
\renewcommand{\thefigure}{A\arabic{figure}}

Hyperparameter optimisation search spaces are provided below. Early Stopping Callback (ESC) patience, number of HPO trials and maximum number of epochs were determined based on experimentation and computational requirements of the model.

\begin{table}[H]
\centering
\caption{LightGBM}
{
\begin{tabular}{|r|c|c|c|}
    \hline
    Hyperparameter & Search space & Selected value (U) & Selected value (A)\\
    \hline
    Number of leaves & $[15..62]$ & 56 & 62 \\
    Number of estimators & $[50..100]$ & 94 & 85 \\
    Subsample for bin & $[100k..200k]$ & 200k & 200k \\
    Minimum child samples & $[10..40]$ & 18 & 31 \\
    Subsample & $[0.1,1.00]$ & 0.89 & 0.76 \\
    Lags & $[24..336]$ & 192 & 72 \\
    Learning rate & $10^{[-5..-1]}$ & $10^{-1}$ & $10^{-1}$ \\
    ESC patience & - & 10 & 10 \\
    Number of trials & - & 30 & 30 \\
    \hline
\end{tabular}
}
\end{table}

\begin{table}[H]
\centering
\caption{DeepAR}
{
\begin{tabular}{|r|c|c|c|}
    \hline
    Hyperparameter & Search space & Selected value (U) & Selected value (A)\\
    \hline
    Hidden dimensions & $[12..50]$ & 22 & 14\\
    Batch size & $[16..64]$ & 35 & 22\\
    Number of RNN layers & $[1..4]$ & 1 & 1\\
    Dropout & $[0.0,1.0]$ & 0.2 & 0.2\\
    Learning rate & $10^{[-5..-1]}$ & $10^{-3}$ & $10^{-4}$\\
    Input chunk length & $[24..336]$ & 192 & 288\\
    Training lentgh & $[124..336]$ & 312 & 24\\
    ESC patience & - & 10 & 10 \\
    Number of trials & - & 80 & 80 \\
    Max epochs & - & 100 & 100 \\
    \hline
\end{tabular}
}
\end{table}

\begin{table}[H]
    \centering
    \caption{TFT}
    {
    \begin{tabular}{|r|c|c|c|}
        \hline
        Hyperparameter & Search space & Selected value (U) & Selected value (A)\\
        \hline
        Hidden size & $[4..16]$ & 16 & 16\\
        LSTM layers & $[1..4]$ & 1 & 1 \\
        Number of attention heads & $[1..4]$ & 4 & 4 \\
        Dropout & $[0.0,1.0]$ & 0.1 & 0.1 \\
        Hidden continuous size & $[4..16]$ & 8 & 8 \\
        Learning rate & $10^{[-5..-1]}$ & $10^{-3}$ & $10^{-3}$\\
        Input chunk length & $[24..336]$ & 168 & 168\\
        Batch size & $[16..64]$ & 32 & 32\\
        ESC patience & - & 3 & 3 \\
        Number of trials & - & 10 & 10 \\
        Max epochs & - & 20 & 20 \\
        \hline
    \end{tabular}
    }
\end{table}

\begin{table}[H]
    \centering
    \caption{N-BEATS}
    {
    \begin{tabular}{|r|c|c|}
        \hline
        Hyperparameter & Search space & Selected value (U) \\
        \hline
        Number of stacks & $[15..60]$ & 37 \\
        Number of blocks & $[1..2]$ & 1 \\
        Number of layers & $[2..8]$ & 3 \\
        Layer widths & $\{128,256,512\}$ & 256 \\
        Batch size & $[16..64]$ & 51 \\
        Learning rate & $10^{[-5..-1]}$ & $10^{-5}$ \\
        Input chunk length & $[24..336]$ & 192 \\
        ESC patience & - & 10  \\
        Number of trials & - & 20 \\
        Max epochs & - & 100 \\
        \hline
    \end{tabular}
    }
\end{table}

\pagebreak

\section{Appendix: Technical analysis indicators}\label{appendix_b}
\setcounter{table}{0}
\renewcommand{\thetable}{B\arabic{table}}
\setcounter{figure}{0}
\renewcommand{\thefigure}{B\arabic{figure}}

Description of all technical analysis indicators used as an input is provided below.

\subsection{Momentum Indicators}
\subsubsection{Absolute oscillator}
Absolute oscillator (AO) (known as absolute price oscillator in econometric context) is a simple comparison between fast and slow moving averages. It is calculated as follows:
    \[ \text{AO} = \text{SMA}_{12}-\text{SMA}_{16} \]
here with arbitrary window sizes.

\subsubsection{Chande momentum oscillator}
Chance momentum oscillator (CMO) aims to capture the momentum of the series $\{y_0, y_1, \dots\}$ by comparing the relative difference between preceding "up values" ($S_u$) and "down values" ($S_d$). It is calculated as follows (in our case with $n = 168$):
    \begin{align*}
        S_u &= \sum_{i=1}^{n} y_{t-i}\mathbb{1}\{y_{t-i} < y_t\} \\
        S_d &= \sum_{i=1}^{n} y_{t-i}\mathbb{1}\{y_{t-i} > y_t\} \\
        \text{CMO} &= 100 \times \frac{S_u-S_d}{S_u+S_d}
    \end{align*}
\subsubsection{Momentum indicator}
Momentum indicator (MOM) is a "naive" momentum indicator: a simple difference between observed value and lagged value:
    \[ \text{MOM} = y_t - y_{t-n} \]
\subsubsection{Percentage oscillator}
Percentage oscillator is a momentum indicator showing the difference between fast and slow exponential moving averages in proportion of the slower one:
    \[ \text{PO} = \frac{\text{SMA}_{12} - \text{SMA}_{26}}{\text{SMA}_{26}} \times 100\]
\subsubsection{Rate of change indicator}
Rate of change indicator (ROC) is MOM proportional to the lagged value:
    \[ \text{ROC} = (y_t-y_{t-n}) / y_{t-n} \]
\subsubsection{Relative Strength Index}
Relative strength index (RSI) is a momentum indicator similiar to CMO:
    \begin{align*}
        \text{UD}_{n} &= \sum_{i=0}^{n} y_i - y_{i-1}\mathbb{1}\{y_i > y_{i-1}\} \\
        \text{DD}_{n} &= \sum_{i=0}^{n} y_{i-1} - y_{i}\mathbb{1}\{y_i < y_{i-1}\} \\
        \text{RSI} &= 100 - 100/\left(1+\frac{\text{UD}_{n}}{\text{DD}_{n}}\right)
    \end{align*}
where $\text{UD}_{n}$ is sum of positive differences and $\text{DD}_{n}$ is sum of negative differences withing the rolling window.
\hspace{1cm}
\subsection{Math Transform}
Math transform class consists of simple mathematical operations performed on each member of the set. For brevity we will define $S = \{ y_{t-i} \}_{i=0}^{n}$.
\subsubsection{Vector Trigonometric Atan $\{\text{ATAN}=\arctan(x) \,:\, x \in S \}$} 
\subsubsection{Vector Trigonometric Cos $\{\text{COS}=\cos(x) \,:\, x \in S \}$}
\subsubsection{Vector Hyperbolic Cos $\{\text{COSH}=\cosh(x) \,:\, x \in S \}$}
\subsubsection{Vector Arithmetic Exp $\{\text{EXP}=\exp(x) \,:\, x \in S \}$}
\subsubsection{Vector Trigonometric Sin $\{\text{SIN}=\sin(x) \,:\, x \in S \}$}
\subsubsection{Vector Hyperbolic Sin $\{\text{SINH}=\sinh(x) \,:\, x \in S \}$}
\subsubsection{Vector Square Root $\{\text{SQRT}=\sqrt{(x)} \,:\, x \in S \}$}
\subsubsection{Vector Trigonometric Tan $\{\text{TAN}=\tan(x) \,:\, x \in S \}$}
\subsubsection{Vector Hyperbolic Tan $\{\text{TANH}=\tanh(x) \,:\, x \in S \}$}
\hspace{1cm}
\subsection{Overlap Studies}
\subsubsection{MidPoint over period}
    \[\text{MIDPOINT} = \frac{\max(S)-\min{S}}{2}\]
\subsubsection{Simple Moving Average}
    \[\text{SMA}= \frac{1}{n} \sum_{i=0}^{n} y_{t-i} \]
\subsubsection{Weighted Moving Average}
    \[ \text{WMA} = \sum_{i=0}^{n}\frac{y_{t-i}n-i}{n-i} \]
\hspace{1cm}
\subsection{Cycle Indicators}
\subsubsection{Kaufman Adaptive Moving Average}
Kaufman adaptive moving average (KAMA) is a filter that aims to reduce signal noice by accounting for volatility. It is calculated as follows:
    \begin{align*}
        C &= | y - y_{t-n} | \\
        V &= \sum_{i=0}^{n}|y_{t-i}-y_{t-i-1}| \\
        \text{ER} &= C / V \\
        \text{SC} &= (\text{ER}\times(\text{SMA}_{2}-\text{SMA}_{30}) + \text{SMA}_{30})^2 \\
        \text{KAMA}_{t} &= \text{KAMA}_{t-1} + \text{SC} (y_t - \text{KAMA}_{t-1})
    \end{align*}

where C=change, V=volatility, ER=efficiency ratio and SC=smoothing constant.
\subsubsection{Triangular Moving Average}
Triangular moving average (TRIMA) is an average of $n$ SMA functions:
    \[\text{TRIMA}= \sum_{i=0}^{n} \text{SMA}_{i} / 168 \]
\subsection{Statistics Functions}
Given the overal form of linear regression:
    \[ y_t = X\beta + \epsilon\]
then by definition:
\subsubsection{Linear Regression Angle $\text{LINEARREGANGLE}=\arctan({\beta})$}
\subsubsection{Linear Regression Intercept $\text{LINEARREGINTERCEPT}=\epsilon$}
\subsubsection{Linear Regression Slope $\text{LINEARREGSLOPE}=\beta$}
\subsubsection{Standard Deviation $\text{STDDEV}=\sigma$}
\subsubsection{Variance $\text{VAR}=V(S)$}
\hspace{1cm}
\subsection{Math Operators}
\subsubsection{Highest value over in window $\text{MAX}=\max(S)$} 
\subsubsection{Index of highest value in window}
    MAXINDEX is the index of the highest observed value of $S$
\subsubsection{Lowest value over a specified period $\text{MIN}=\min(S)$}
\subsubsection{Index of lowest value over a specified period}
    MININDEX is the index of the highest observed value of $S$ 
\subsubsection{Summation $\text{SUM}=\Sigma S$}

\pagebreak

\section{Appendix: Geographic locations of external data sources}\label{appendix_c}

GPS coordinates of the traffic monitoring nodes and health district hospitals along with their geometric distance to study hospital are provided below.

\setcounter{table}{0}
\renewcommand{\thetable}{C\arabic{table}}
\setcounter{figure}{0}
\renewcommand{\thefigure}{C\arabic{figure}}

\begin{table}[H]
\centering
\caption{Locations of automatic traffic monitoring nodes}
\begin{tabular}{rcccc}
\toprule
{} &  Identifier &   Latitude &  Longitude &  Distance (km) \\
\midrule
0  &         203 &  61.585140 &  23.429460 &             22 \\
1  &         204 &  61.912381 &  22.975875 &             63 \\
2  &         210 &  61.315328 &  22.838086 &             56 \\
3  &         401 &  61.408751 &  23.765753 &             11 \\
4  &         402 &  61.149148 &  24.057586 &             42 \\
5  &         404 &  61.549823 &  24.071581 &             15 \\
6  &         406 &  61.503859 &  23.368056 &             24 \\
7  &         409 &  62.013430 &  24.027612 &             58 \\
8  &         421 &  61.544140 &  23.607493 &             12 \\
9  &         422 &  61.241797 &  23.738401 &             30 \\
10 &         431 &  61.478619 &  23.977413 &              9 \\
11 &         433 &  61.285640 &  23.955748 &             26 \\
12 &         435 &  61.463652 &  23.808471 &              5 \\
13 &         436 &  61.478807 &  23.586782 &             12 \\
14 &         438 &  61.508408 &  23.697679 &              6 \\
15 &         439 &  61.504163 &  23.821503 &              0 \\
16 &         440 &  61.252700 &  23.837494 &             28 \\
17 &         443 &  61.504828 &  23.711674 &              5 \\
18 &         445 &  61.115503 &  23.647575 &             44 \\
19 &         448 &  61.731480 &  24.708950 &             54 \\
20 &         449 &  61.460262 &  23.745350 &              6 \\
21 &         450 &  61.515739 &  23.568131 &             13 \\
22 &         451 &  61.528153 &  23.939764 &              7 \\
23 &         452 &  61.503156 &  23.724675 &              5 \\
24 &         453 &  61.502302 &  23.739676 &              4 \\
25 &         455 &  61.517417 &  23.654093 &              9 \\
26 &         456 &  61.511616 &  23.679207 &              7 \\
27 &         457 &  61.506480 &  23.785048 &              2 \\
28 &         458 &  61.504198 &  23.852788 &              2 \\
29 &         460 &  61.452590 &  23.636382 &             11 \\
30 &         462 &  61.477796 &  23.588359 &             12 \\
31 &         463 &  61.459451 &  23.617727 &             12 \\
32 &         464 &  61.479608 &  23.775252 &              4 \\
33 &         471 &  61.439193 &  23.759727 &              8 \\
34 &         921 &  61.488543 &  25.146808 &             71 \\
\bottomrule
\end{tabular}
\end{table}

\begin{table}[H]
\centering
\caption{Locations of follow-up care hospitals}
\begin{tabular}{rcccc}
\toprule
{} & Identifier &   Latitude &  Longitude &  Distance (km) \\
\midrule
0  &       HC01 &  61.284161 &  24.034778 &             27 \\
1  &       HC02 &  61.637557 &  23.193016 &             36 \\
2  &       HC03 &  61.769611 &  23.073668 &             49 \\
3  &       HC04 &  61.458140 &  24.084322 &             15 \\
4  &       HC05 &  61.317678 &  23.747963 &             21 \\
5  &       HC06 &  61.480554 &  23.465854 &             19 \\
6  &       HC08 &  61.663521 &  24.365686 &             34 \\
7  &       HC09 &  61.461408 &  23.637711 &             11 \\
8  &       HC10 &  61.982704 &  24.076372 &             55 \\
9  &       HC11 &  61.337111 &  22.923229 &             51 \\
10 &       HC12 &  61.082570 &  23.561754 &             49 \\
11 &       HC13 &  61.271148 &  24.030153 &             28 \\
12 &       HC14 &  62.240749 &  23.759871 &             82 \\
13 &       HC15 &  61.546931 &  23.600148 &             12 \\
14 &        RHA &  61.484384 &  23.757655 &              4 \\
15 &        RHB &  61.270092 &  24.031116 &             29 \\
\bottomrule
\end{tabular}
\end{table}

\pagebreak

\section{Appendix: Computational cost}\label{appendix_d}
\setcounter{table}{0}
\renewcommand{\thetable}{D\arabic{table}}
\setcounter{figure}{0}
\renewcommand{\thefigure}{D\arabic{figure}}

All analysis were performed using Atos Bullsequana X400 supercomputing cluster provided by IT Center for Science (CSC). Total runtimes for each model including hyperparameter tuning, initial parameter estimation and backtesting are provided in Table \ref{tab:runtime}.

\begin{table}[H]
\centering
\caption{Computational cost measured in walltime for hyperparameter optimisation (HPO),        parameter estimation and backtesting in minutes. CPU=Intel Xeon Cascade Lake 2,1 GHz,        GPU1=Nvidia Volta V100, GPU2=Nvidia Ampere A100}
\label{tab:runtime}
\begin{tabular}{rccccc}
\toprule
{} &      HPO &     Fit &  Backtest &    Total & Accelerator \\
Name       &          &         &           &          &             \\
\midrule
SN-U       &        - &       - &      0.05 &     0.10 &         CPU \\
HWAM-U     &        - &       - &      0.72 &     0.75 &         CPU \\
HWDM-U     &        - &       - &      0.85 &     0.88 &         CPU \\
LightGBM-U &     4.00 &    0.27 &      0.23 &     4.52 &         CPU \\
TFT-U      &    25.33 &    2.65 &      0.88 &    28.90 &        GPU2 \\
DeepAR-U   &   142.93 &    1.50 &      0.77 &   145.25 &        GPU1 \\
ARIMA-U    &        - &       - &    196.47 &   196.50 &         CPU \\
N-BEATS-U  &   221.72 &   11.67 &      0.93 &   234.33 &         CPU \\
LightGBM-A &   287.85 &   31.22 &      0.78 &   319.88 &        GPU1 \\
DeepAR-A   &   327.55 &    1.00 &      1.03 &   329.62 &        GPU1 \\
TFT-A      &   346.02 &   63.45 &     12.38 &   421.90 &        GPU2 \\
Total      &  1355.40 &  111.75 &    215.10 &  1682.63 &           - \\
\bottomrule
\end{tabular}
\end{table}

\end{document}